\documentclass{article}

\usepackage[T1]{fontenc}

\usepackage{graphicx}
\usepackage{hyperref}
\hypersetup{
  setpagesize  = false,
  colorlinks   = true,    
  urlcolor     = blue,    
  linkcolor    = black,   
  citecolor    = black    
}
\usepackage{authblk}
\usepackage{geometry}
\usepackage{setspace}
\usepackage{verbatim}
\usepackage{amsmath}
\usepackage{amssymb}
\usepackage{gensymb}
\usepackage{amsfonts}

\usepackage[sorting=none]{biblatex}
\title {Segmentation by Factorization: Unsupervised Semantic Segmentation for Pathology by Factorizing Foundation Model Features }
\author{
 Jacob Gildenblat$^\ast$, Ofir Hadar$^\ast$}
\affil{DeePathology.ai}
\date{}

\usepackage{fancyhdr}
\begin{document}

\maketitle
\vspace{6pt}

\def\thefootnote{$^\ast$}\footnotetext{These authors contributed equally to this work}

\begin{abstract}

We introduce Segmentation by Factorization (F-SEG), an unsupervised segmentation method for pathology that generates segmentation masks from pre-trained deep learning models. F-SEG allows the use of pre-trained deep neural networks, including recently developed pathology foundation models, for semantic segmentation. It achieves this without requiring additional training or fine-tuning, by factorizing the spatial features extracted by the models into segmentation masks and their associated concept features. We create generic tissue phenotypes for H\&E images by training clustering models for multiple numbers of clusters on features extracted from several deep learning models on TCGA \cite{tcga_dataset}, and then show how the clusters can be used for factorizing corresponding segmentation masks using off-the-shelf deep learning models. Our results show that F-SEG provides robust unsupervised segmentation capabilities for H\&E pathology images, and that the segmentation quality is greatly improved by utilizing pathology foundation models. We discuss and propose methods for evaluating the performance of unsupervised segmentation in pathology. 
\end{abstract}

\vspace{6pt}

\section{Introduction}
Segmentation models are widely used in applications of digital pathology for focusing on regions of interest, such as detecting tumor regions, or performing a subsequent analysis only inside specific regions, or excluding non-cellular areas \cite{xu2015deep, xu2016deep}. These segmentation masks can also serve as predictive features by quantifying the area of a specific region or counting cells within that region (e.g., {\cite{diao2021human}). We explore the use of existing pre-trained neural networks for semantic segmentation, without the need to train dedicated segmentation models for new datasets. As research progresses in developing pre-trained models for pathology, often referred to as foundation models when these are large models trained on large datasets, leveraging these off-the-shelf models can reduce barriers to create segmentation models and potentially enhance segmentation accuracy as the pre-trained models improve.

This capability is useful in various tasks, such as expediting semantic segmentation annotations through automatic annotations, proposing regions of interests in slides, quantifying objects in different region types, or enabling interpretability by examining how model outputs interact with different region types.

Our contributions are as follows:

\begin{enumerate}
    \item We demonstrate that Non-Negative Matrix Factorization (NMF) \cite{NIPS2005_d58e2f07, tandon2010sparse} on top of spatial activations extracted from deep neural networks can be used to generate consistent and meaningful semantic segmentation masks for pathology. We propose two ways of achieving this. One is by applying NMF on top of the spatial activations to factorize them into a segmentation mask and concept features, and then classifying the concept features with a clustering model that can be created on a different dataset. The second way is to apply NMF where one of the matrices is fixed and set to be the cluster centers, and then solve for a segmentation mask that corresponds to these clusters. The achieved segmentation does not require training deep networks, allows utilizing off-the-shelf deep learning models, and allows segmentation into a varying number of categories depending on the need.
    \item We use this to create a generic unsupervised semantic segmentation method for H\&E for a varying number of categories and for several deep learning models, and benchmark the proposed unsupervised segmentation method on H\&E segmentation tasks to show that the achieved segmentation correspond with annotated categories in semantic segmentation datasets. Figure \ref{fig:uni_luad_example} illustrates an example of the output segmentation.
\end{enumerate}

\begin{figure}
\caption{An example of F-SEG semantic segmentation with the UNI foundation model and k=64 TCGA clusters}
\includegraphics[scale=0.26]{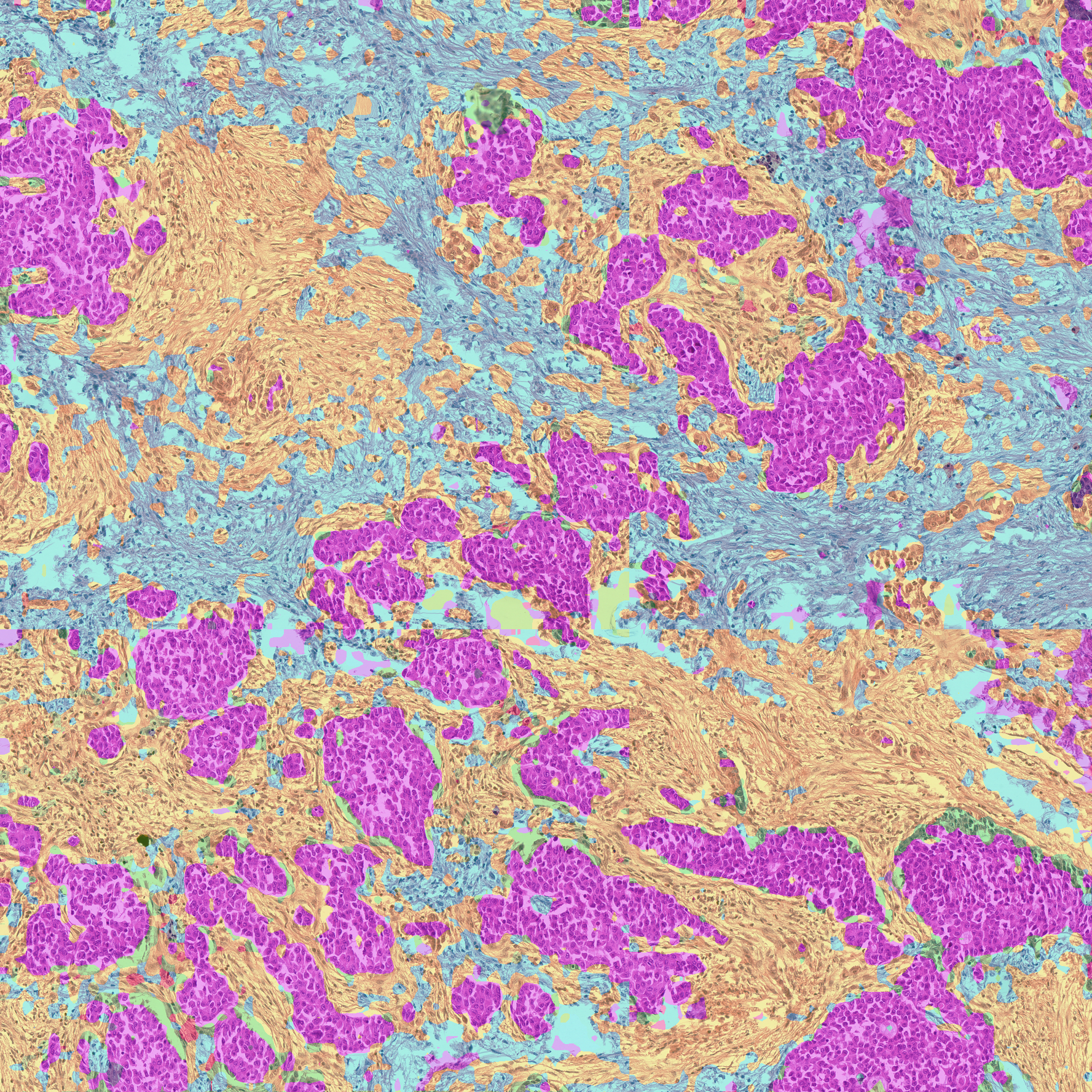}
\label{fig:uni_luad_example}
\end{figure}

\section{Related work}
Image segmentation has been a central topic in computer vision research for decades. It is typically categorized into three types: semantic segmentation \cite{asgari2021deep}, instance segmentation \cite{minaee2021image}, and panoptic segmentation \cite{kirillov2019panoptic}. This paper focuses on semantic segmentation, where the objective is to classify each pixel in an image into one of several categories. In supervised semantic segmentation, the process involves images with known categories and detailed pixel-wise annotations. Deep learning models like U-Net \cite{ronneberger2015u} have achieved impressive results in this area and are commonly used across various domains, including medical imaging \cite{azad2022medical}.

In contrast, unsupervised semantic segmentation does not rely on predefined categories or pixel-wise annotations. Instead, the goal is to discover representative categories from the data itself and then generate pixel-wise classifications based on these categories. Previous approaches often involve designing a specific loss function for unsupervised segmentation and training a deep learning model such as U-Net using this loss. For example, \cite{wnet} introduces an encoder/decoder framework where the encoder performs a k-way segmentation into k unknown categories, and the decoder reconstructs the image from features produced by the encoder. They employ a soft normalized cut loss to ensure similarity within the same category and dissimilarity between different categories. On the other hand, \cite{useg2020} uses a loss function applied to the output of a U-Net decoder to ensure that pixels with the same label have similar features, ensuring spatial continuity while promoting a larger amount of clusters.

Weakly supervised semantic segmentation, another related approach \cite{wei2016stc, huang2018weakly}, involves the use of weaker labels such as bounding boxes or image-level labels rather than pixel-wise annotations. Methods like Grad-CAM \cite{GradCAM} use these weaker labels to generate attribution maps, which are then processed to create pseudo-segmentation masks for training purposes. Affinity-Net \cite{Ahn_2018_CVPR} is one such method that leverages these pseudo-masks for training. Our approach shares similarities with this method in that it uses activations from a model to create pseudo-segmentation masks, but unlike these methods, we do not use any predefined image labels.

Deep Feature Factorization \cite{collins2018deep} applies NMF on the activations of deep neural networks to visualize similar features in different parts of an image and perform co-segmentation \cite{rother2006cosegmentation, mukherjee2009half} across multiple images. Unlike methods that rely on attribution maps, this approach does not require tile labels. Inspired by Deep Feature Factorization, our method aims to utilize the features of a pretrained network based solely on its activations, without the need for annotations.

\section{Methods}

\subsection{Factorizing spatial features of deep neural networks into concepts and segmentation masks}
Following the approach proposed in Deep Feature Factorization \cite{collins2018deep}, we apply NMF on the spatial activations of a pre-trained neural network on an image. The goal is to factorize these activations into a concept feature matrix and a semantic segmentation of the concepts. We start by extracting the spatial activations of a pre-trained network, assumed here to be non negative, resulting in a tensor of shape $rows \times cols \times channels$. In the case of a vision transformer model, each token in the output origins from an input location in the image, so we ignore the class token. We then reshape this tensor into a matrix $A$ of shape  $(rows \times cols) \times channels$ and then factorize $A$ with NMF into $k$ components:
    $$A_{(rows \times cols) \times channels} \approx W_{(rows \times cols) \times k} \times H_{k \times channels}$$

In this factorization, $W$ represents a per-pixel contribution matrix, indicating the contribution of every pixel to the $k$ concepts.
For simplicity, we refer to $W_{ijm}$ as the contribution of the pixel $ij$ to the $m$'th concept. 
$H$ represents a concept feature matrix, storing the $k$ representative feature vectors of the concepts.

We assign each pixel a concept $\beta_{ij}$ by identifying the concept with the largest contribution to that pixel:
 
$$\beta_{ij} = argmax_{m}W_{ijm}$$

\subsection{Methods for achieving consistent concepts between images}
The factorized concepts in $\beta$ might have different meanings and ordering in different images. For example, a factorized concept corresponding to "Tumor" might be the first concept in one image, but the third in another. To achieve consistent semantic segmentation across images, we map these concepts to a common set of labels. This can be done using any method that assigns feature vectors into categories, for example by utilizing existing classification models. We achieve this by fitting a k-means clustering model \cite{hartigan1979algorithm} on 1D features extracted from neural networks by taking the spatial average of the features from the layer of choice in the neural network.

Finally, at the end of the pipeline, the factorized segmentation mask is resized to match the shape of the input image. We next propose two methods for utilizing the clustering model centers for consistent concepts. 

\subsubsection{Assigning factorized concepts to similar clusters}

For each pixel $i,j$ the semantic segmentation category $\hat{y}_{ij}$ is selected by taking the cluster center with the highest cosine similarity to the factorized concept feature at that pixel: 
$$\hat{y_{ij}} = \operatorname*{argmax}_{k} \frac{\mathbf{\mu}_k \cdot \mathbf{H_{\beta_{ij}}}}{\|\mathbf{\mu}_k\| \ \|\mathbf{H_{\beta_{ij}}}\|}$$

Where $\mathbf{\mu}_k$ is the k'th clustering center, and $\mathbf{H_{\beta_{ij}}}$ is the concept feature vector from the concept feature matrix $H$ 

The pipeline for F-SEG when classifying factorized concepts into the most similar cluster is illustrated in figure \ref{fig:pipeline_cosine_similarity}.

\begin{figure}[t]
\caption{Overview of F-SEG when classifying factorized concepts into the most similar cluster}
\includegraphics[scale=0.12]{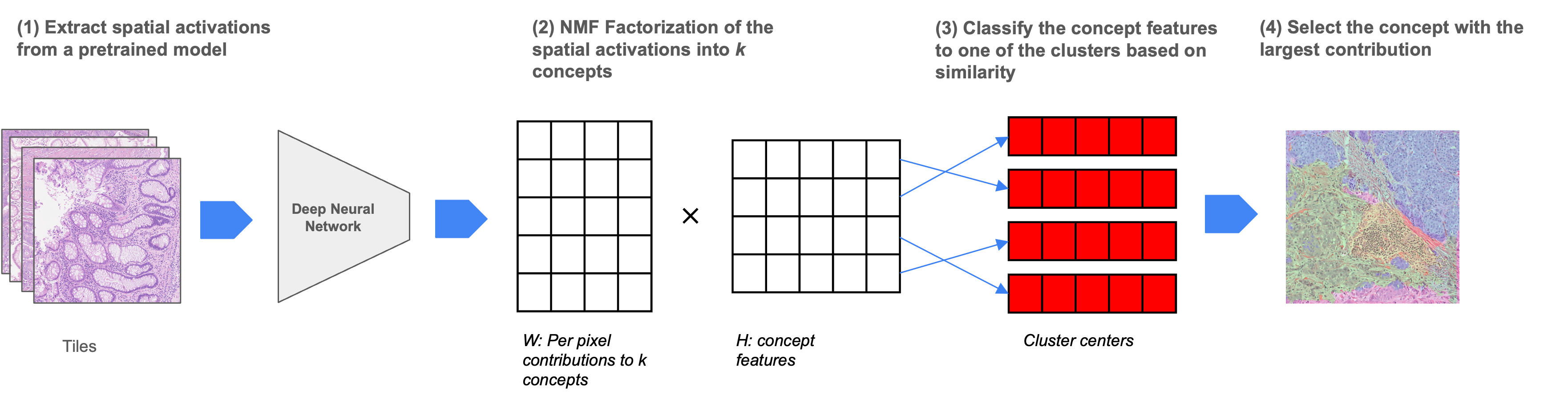}
\label{fig:pipeline_cosine_similarity}
\end{figure}

\subsubsection{Solving for segmentation masks that correspond to cluster centers}

Here instead of applying NMF to solve for both the $H$ and $W$ matrices, we set the matrix $H$ to be equal to the cluster centers, and solve only for $W$. This finds a segmentation mask that corresponds to these clusters. 

The pipeline for F-SEG when fixing the concept matrix is illustrated in figure \ref{fig:pipeline_fixed_h}.

\begin{figure}[t]
\caption{Overview of F-SEG when fixing the concept matrix to be cluster centers}
\includegraphics[scale=0.12]{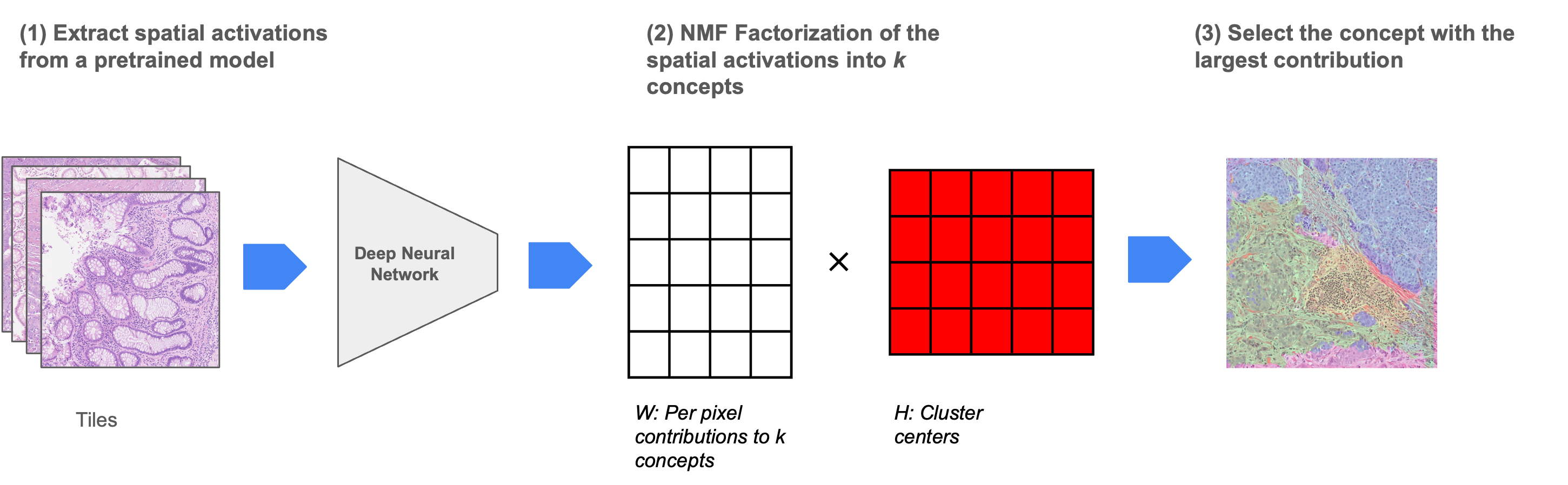}
\label{fig:pipeline_fixed_h}
\end{figure}

\subsection{Unsupervised semantic segmentation for H\&E images}

Here we apply F-SEG for unsupervised semantic segmentation of H\&E histology images. To create clustering models used by the factorization, we extract image tiles of size $256 \times 256$ from whole-slide images (WSIs) at a magnification of 10x from the TCGA dataset. The tiles are randomly sampled, with 200 tiles taken per slide from a total of 11,000 slides, covering a wide variety of tissue types and morphologies.

For clustering we use k-means with varying values of $k$ (16, 32, 64, 128, and 256) allowing segmentation into different numbers of categories. By adjusting the k-value, we can classify the image regions into varying levels of granularity, allowing the segmentation to capture either broad tissue regions or finer anatomical structures, allowing a quick adaptation of the segmentation to different requirements.

We compare F-SEG with three deep learning models: the Resnet50 model \cite{he2016deep}, and two recent foundation models, the UNI model \cite{chen2024uni} and the Prov-GigaPath model \cite{xu2024gigapath}, both of which are based on vision transformers \cite{dosovitskiy2020vit}. The Resnet50 model serves as our baseline, having been pre-trained on the ImageNet dataset, which is commonly used for general image classification tasks. In contrast, the UNI and Prov-GigaPath models are state-of-the-art vision transformers, designed specifically for large-scale image analysis and particularly suited to extracting complex spatial relationships in digital pathology images.

For feature extraction using Resnet50, we observed through early experimentation that better clustering performance was achieved when using the activations from the second-to-last Resnet block, that has a higher spatial resolution at it's output, rather than the final block.
For the foundation vision transformer models, we use the activations from the final self-attention block. However, since the outputs of these vision transformer models can contain negative values, in contrast to the Resnet50 model, which includes a Rectified Linear Unit (ReLU) activation \cite{agarap2018deep} to zero out negative values, we apply a ReLU to these features as a post-processing step. This ensures that there are no negative values in the activations factorized with Non Negative Matrix Factorization. However this is the most simple choice, and other ways could be potentially explored.

Finally we apply a Global Average Pooling (GAP) layer on top of the extracted features, and remove their spatial dimensions, resulting in 1D features suitable for clustering with k-means.

\subsection{Evaluation Datasets}
\subsubsection{Breast Cancer Semantic Segmentation}

The Breast Cancer Semantic Segmentation (BCSS) dataset \cite{amgad2019structured} contains over 20,000 segmentation annotations of tissue regions from breast cancer images sourced from The Cancer Genome Atlas (TCGA). 
The dataset is divided into training and test sets, with pixel-wise category labels for both sets. In the training set, the Tumor category constitutes 15.17\% of the pixels, the Stroma category 34.89\%, the Inflammatory category 30.37\%, the Necrosis category 9.47\%, and the Other category 5.86\%.

The test set shows a different distribution, with the Tumor category making up 33.55\% of the pixels, the Stroma category 24.3\%, the Inflammatory category 26.24\%, the Necrosis category 9.23\%, and the Other category 3.26\%. The variation in pixel distribution between the training and test sets provides a robust basis for evaluating the performance and generalization capabilities of segmentation models.

\subsubsection{WSSS4LUAD}

The Weakly-supervised Tissue Semantic Segmentation
for Lung Adenocarcinoma (WSSS4LUAD) challenge \cite{graham2021lizard} aims to perform tissue semantic segmentation in H\&E stained Whole Slide Images (WSIs) for lung adenocarcinoma. 

For the validation set, a total of 40 patches were manually cropped by the label review board. This set included 9 large patches (approximately $1500-5000$ \(\times\) $1500-5000$ pixels) and 31 small patches (approximately $200-500$ \(\times\) $200-500$ pixels). For the test set, a total of 80 patches were manually cropped, consisting of 14 large patches (approximately $1500-5000$ \(\times\) $1500-5000$ pixels) and 66 small patches (approximately $200-500$ \(\times\) $200-500$ pixels). 

The distribution of pixel labels across different categories in the validation set is 42.18\% pixels for the Tumor category, 27.78\% pixels for the Stroma category, 26.63\% pixels for the Background category, and 3.39\% pixels for the Normal category. In the test set, the distribution is 46.25\% pixels for the Tumor category, 33.63\% pixels for the Stroma category, 17.66\% pixels for the Background category, and 2.44\% pixels for the Normal category.

Since the training set is labeled weakly with only image-level annotations, we focused our evaluation on the validation and test sets, which have precise pixel-level annotations.

\subsection{Evaluation methods}
In this section, we outline the evaluation methods used to assess the performance of our methods. We propose two evaluation methods for unsupervised segmentation. First, a "linear probing" method specific to F-SEG that benchmarks the potential of the factorized features and segmentation masks, where a linear classifier on top of the factorized features is used to classify them into ground truth categories, evaluating both the factorized features, and the segmentation quality. And a second matching-based approach, suitable for unsupervised segmentation methods in general, that benchmarks the unsupervised segmentation by matching the predicted unsupervised categories with the ground truth categories.
\subsubsection{Linear probing}

In this evaluation method we replaced the clustering model with a linear classifier to classify concept features into specific categories. The classifier is trained on top of factorized features that are found to be co-occuring with ground truth categories.To train the classifier, we followed these steps:

1. \textbf{Concept-Category Association:} For each tile, we identified concepts obtained with NMF that were highly associated with certain categories. For each concept \( m \), we calculated the percentage of pixels corresponding to that concept which fell into a ground truth category \( n \).

2. \textbf{Thresholding:} If the percentage of pixels for concept \( m \) in category \( n \) exceeded a predefined threshold, we assigned category \( n \) to that concept. We then recorded the concept feautures \( H_{m} \) along with its label \( n \).

3. \textbf{Training the Linear Classifier:} We used the generated pairs of concept features and their corresponding labels to train a linear classifier \( w \).

4. \textbf{Generating Semantic Segmentation Labels:} Finally, we utilized the trained linear classifier to classify the concept features, resulting in the semantic segmentation labels \( \hat{y} \).
$$\hat{y_{ij}} =  \operatorname*{argmax}_{c=1..n} {w_c\cdot{H_{\beta_{ij}}}}$$

\subsubsection{Matching based on cluster-category frequencies}
In this section, we describe the process of matching cluster indices with ground truth categories. Each cluster is associated with a single ground truth category, though a ground truth category can correspond to multiple clusters.

- Matching Clusters to Ground Truth Categories: For each cluster, we calculate the frequency of pixels from each ground truth category that fall within the cluster. We then match each cluster to the ground truth category that has the highest frequency of pixels within that cluster.

- Handling Unbalanced Categories: The distribution of ground truth categories can be uneven, with some categories being very common and others quite rare. To address this imbalance, we provide an alternative matching method. In this method, we normalize the frequency of pixels in each cluster by the total number of pixels in the respective ground truth category. This normalization allows for matching clusters with ground truth categories when there is a high unbalance in the number of pixels belonging to different ground truth categories. 

\section{Results}

The results for all models and evaluation methods are shown in Figures \ref{fig:results_bcss} and \ref{fig:results_luad}. The foundation models demonstrate a significant improvement over the baseline Resnet50 model pretrained on ImageNet. Specifically, our experiments show that the foundation models consistently outperform Resnet50 across all tested scenarios, achieving better performance in terms of F1 score, and overall robustness.
Fixing the concept matrix $H$ leads to improved performance in all settings, compared to performing full NMF and classifying the factorized concepts.

\begin{figure}
\caption{Evaluating Segmentation by Factorization on the Breast Cancer Semantic Segmentation Dataset}
\includegraphics[scale=0.5]{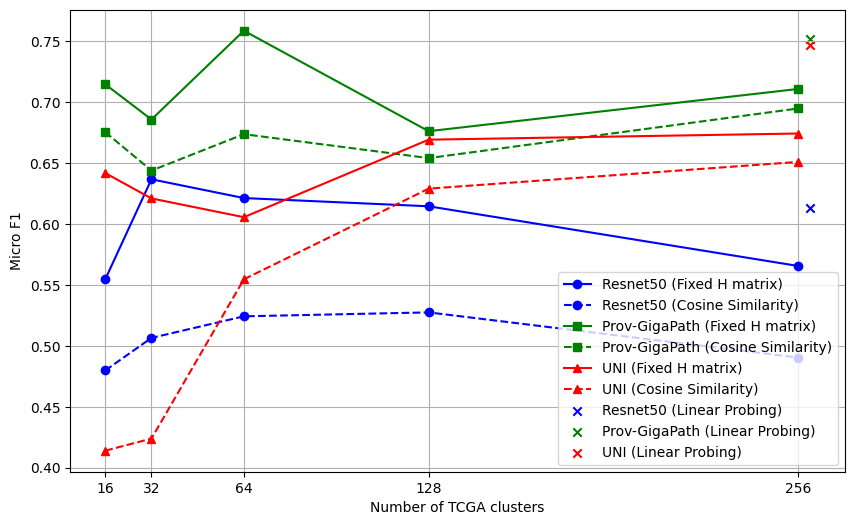}
\label{fig:results_bcss}
\end{figure}

\begin{figure}
\caption{Evaluating Segmentation by Factorization on the WSSS4LUAD Dataset}
\includegraphics[scale=0.5]{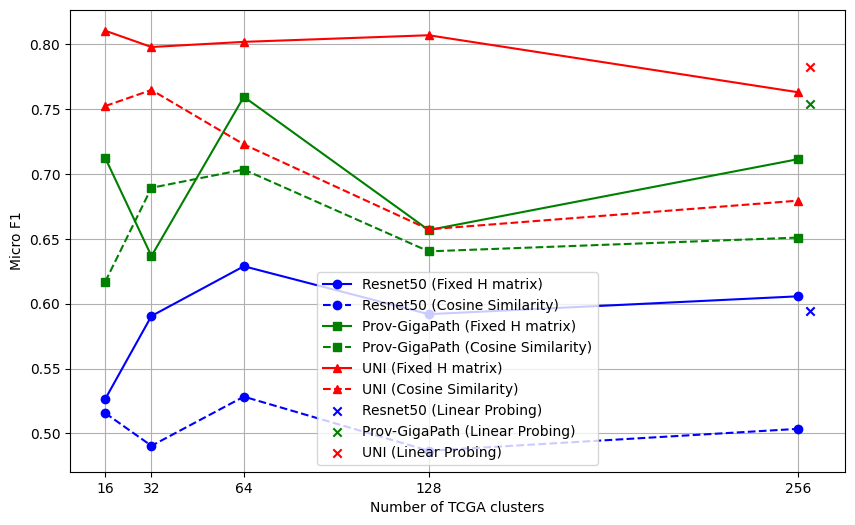}
\label{fig:results_luad}
\end{figure}
\section{Discussion}
We demonstrated how pre-trained deep learning models can be used for unsupervised semantic segmentation, without being required to further train or fine time them. This method allows leveraging large foundation models developed for pathology, and can be used to segment pathology images without any performing any annotations, or without clear definitions of categories. We show that the unsupervised segmentation correspond with meaningful tissue types on semantic datasets datasets.

\printbibliography

\clearpage

\appendix
\section{Examples of F-SEG unsupervised semantic segmentation}

\begin{figure}[!htb]
\caption{An example of F-SEG semantic segmentation with the Prov-GigaPath foundation model and k=16 TCGA clusters}
\includegraphics[scale=0.2]{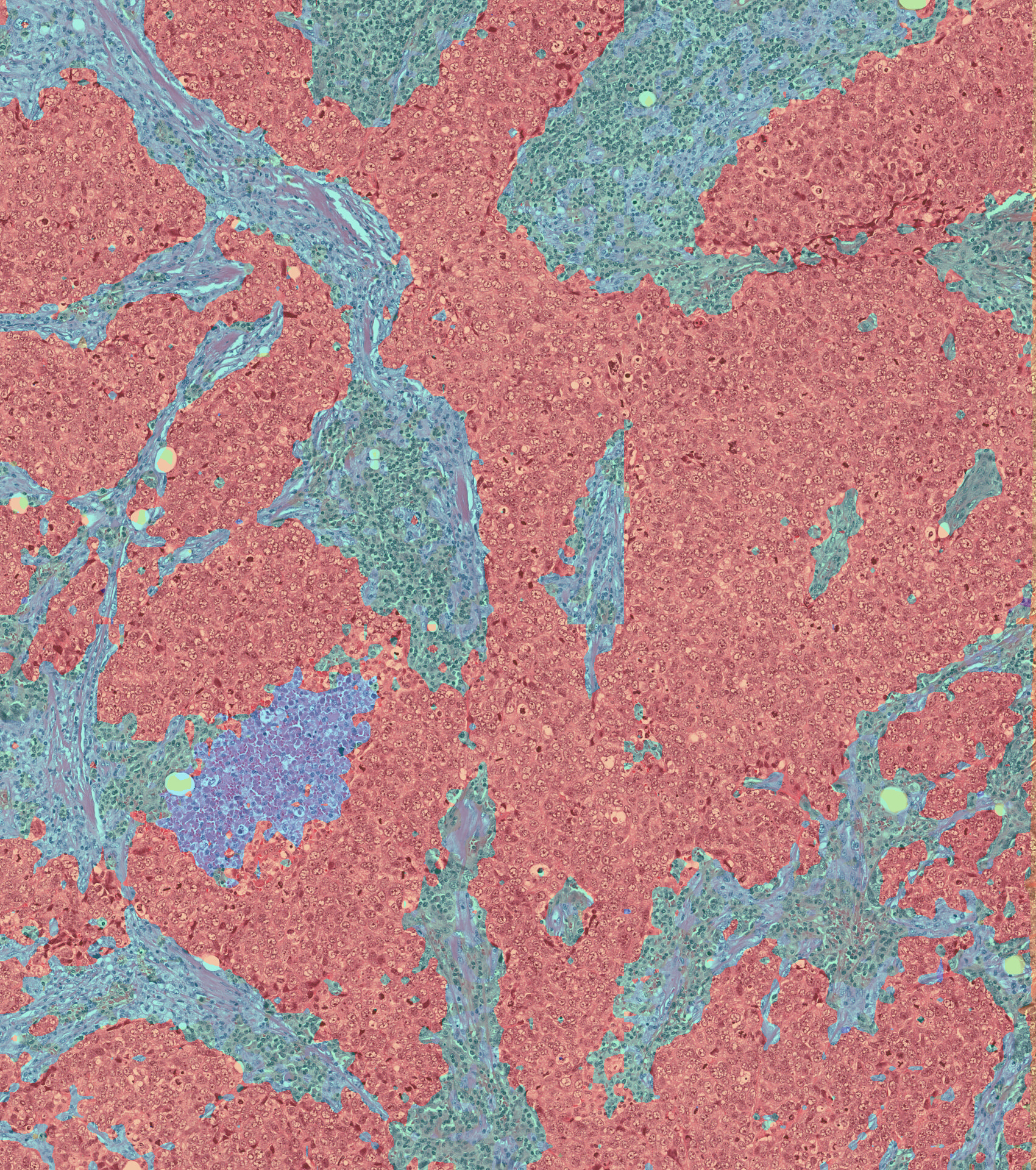}
\label{fig:gigapath_bcss_16_example}
\end{figure}

\begin{figure}
\caption{An example of F-SEG semantic segmentation with the Prov-GigaPath foundation model and k=16 TCGA clusters}
\includegraphics[scale=0.2]{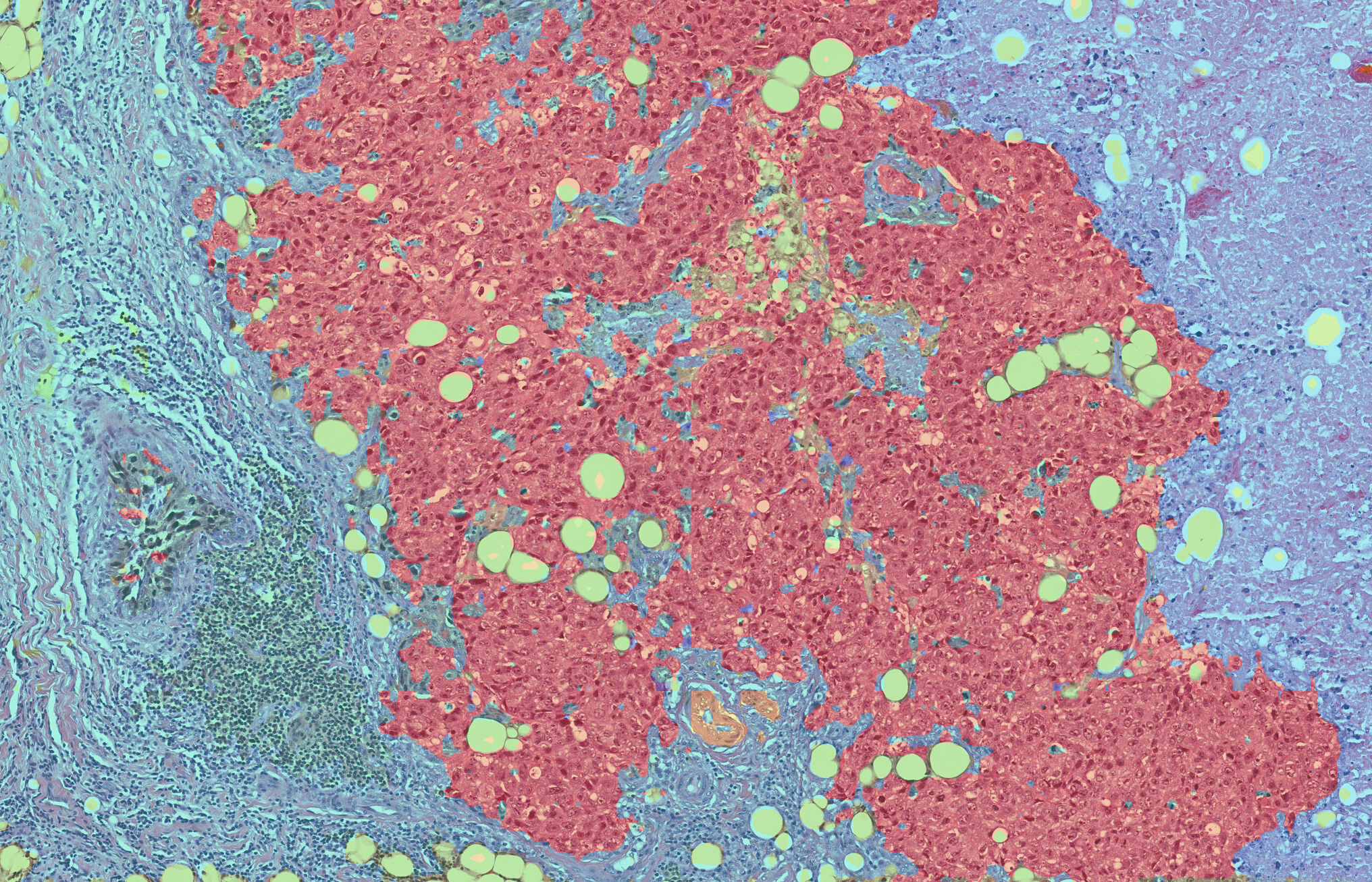}
\label{fig:gigapath_bcss_16_example2}
\end{figure}

\begin{figure}
\caption{An example of F-SEG semantic segmentation with the Prov-GigaPath foundation model and k=64 TCGA clusters}
\includegraphics[scale=0.2]{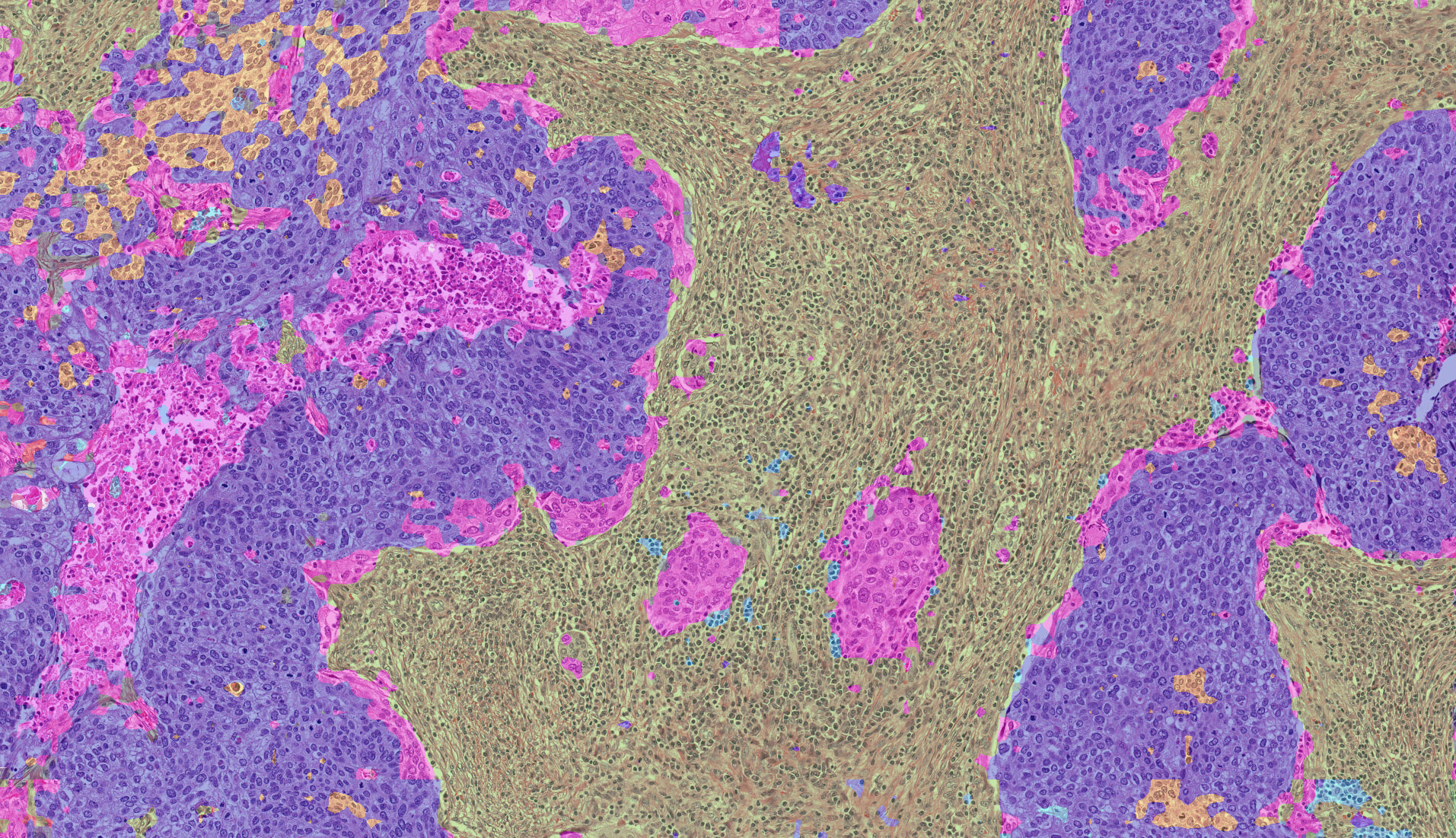}
\label{fig:gigapath_bcss_64_example}
\end{figure}

\end{document}